\documentclass{elsart}
\usepackage{amssymb}
\usepackage{graphicx,epsfig,color,wrapfig}
\usepackage{url,amssymb}

\begin{document}
\begin{frontmatter}

\title{A family of statistical symmetric divergences based on Jensen's inequality}

\author[label1,label2]{Frank Nielsen\corauthref{corr1}}
\corauth[corr1]{Corresponding author. Fax $\#$ (+81) 3-5448-4380.}
\ead{Frank.Nielsen@acm.org}
\ead[url]{http://www.sonycsl.co.jp/person/nielsen/}
\address[label1]{Ecole Polytechnique\\ Computer Science Department (LIX) \\ Palaiseau, France.}
\address[label2]{Sony Computer Science Laboratories, Inc. (FRL),\\
3-14-13 Higashi Gotanda 3F, Shinagawa-Ku,\\
Tokyo 141-0022, Japan.}

\date{September 2010, revised November/December 2011}

\begin{abstract}
We introduce a novel parametric family of symmetric information-theoretic distances based on Jensen's inequality for a convex functional generator. 
In particular, this family unifies the celebrated Jeffreys divergence with the Jensen-Shannon divergence when the Shannon entropy generator is chosen.
We then design a generic algorithm to compute the unique centroid defined as the minimum average divergence.
This yields a smooth family of centroids linking the Jeffreys to the Jensen-Shannon centroid.
Finally, we report on our experimental results.
\end{abstract}


\end{frontmatter}

\def\eH{\mathrm{eH}}
\def\sB{\mathrm{sB}}
\def\sBR{\mathrm{sBR}}
\def\BR{\mathrm{BR}}
\def\sJ{\mathrm{sJ}}
\def\sj{\mathrm{sj}}
\def\dx{\mathrm{d}x}
\def\KL{\mathrm{KL}}
\def\sKL{\mathrm{sKL}}
\def\eKL{\mathrm{eKL}}
\def\JS{\mathrm{JS}}
\def\B{\mathrm{B}}
\def\innerproduct#1#2{ \langle #1, #2\rangle}
\def\argmin{\mathrm{argmin}}

\section{Introduction to statistical distances}
The Shannon differential entropy~\cite{ct-1991} of a continuous probability distribution\footnote{For sake of simplicity and without loss of generality, we consider the probability density function $p$ of a continuous random variable $X\sim p$.
For multivariate densities $p$, the integral notation $\int$ denote the corresponding multi-dimensional integral, so that we write for short $\int p(x)\dx =1$. 
Our results hold for probability mass functions, and probability measures in general.} $p$ measures the amount of uncertainty:

\begin{equation}
H(p)=\int p(x)\log \frac{1}{p(x)} \dx = -\int p(x)\log p(x)\dx.
\end{equation}

The cross-entropy~\cite{ct-1991} measures the amount of extra bits required to compute a code based on an observed empirical probability $\tilde p$ instead of the true probability $p$ (hidden by nature):

\begin{equation}
H(p:\tilde p) =\int p(x)  \log \frac{1}{\tilde p(x)} \dx = -\int p(x)  \log {\tilde p(x)} \dx.
\end{equation}
The ``:'' notation emphasizes on the {\it oriented} aspect~\cite{ct-1991} of the functional: $H(p:q)\not = H(q:p)$. 
The Kullback-Leibler divergence~\cite{kullbackleibler-1951,ct-1991} is a {\it statistical distance measure} computing the {\it relative entropy} as follows:

\begin{eqnarray}
\KL(p:q) & = & \int p(x)\log \frac{p(x)}{q(x)} \dx \\
 & = & H(p:q)- H(p) \geq 0,
\end{eqnarray}
This last inequality is called Gibb's inequality~\cite{ct-1991}, with equality if and only if $p=q$.
We have $H(p:q)=H(p)+\KL(p:q)$.
The Kullback-Leibler divergence can be extended to unnormalized positive distributions (or positive arrays in discrete cases) as follows:
\begin{eqnarray}
\eKL(p:q) &=&  \int \left( p(x)\log \frac{p(x)}{q(x)}+   q(x)-p(x) \right)  \dx,\\
&=& \eH(p:q)- \eH(p) \geq 0,
\end{eqnarray}
with $\eH(p:q)=\int (p(x)\log\frac{1}{q(x)} + q(x))\dx $ and $\eH(p)=\eH(p,p)$.

(R\'enyi based on an axiomatic approach~\cite{Renyi-1961} derived yet another expression for the Kullback-Leibler divergence of unnormalized generalized distributions.)

Many applications in Information Retrieval~\cite{featuresIR-2008,diveval:2001} (IR) requires to deal with a {\it symmetric} distortion measure.
Jeffreys divergence~\cite{Jeffreys-1973} (also called $J$-divergence) symmetrizes the oriented Kullback-Leibler divergence as follows:

\begin{eqnarray}
J(p,q) & = & \KL(p : q)+\KL(q : p) = J(q,p) \label{eq:symJ} \\
 & = & H(p:q)+H(q:p) - (H(p)+H(q)),\\
  & = & \int (p(x)-q(x))\log \frac{p(x)}{q(x)} \dx.
\end{eqnarray}
Here, we replaced ``:'' by ``,'' in the distortion measure to emphasize the symmetric property: $J(p,q)=J(q,p)$.
Jeffreys divergence is interpreted as {\it twice the  average of the cross-entropies minus the average of the entropies}.
One of the drawbacks of Jeffreys divergence is that it may be {\it unbounded} and therefore numerically quite unstable to compute in practice:
For example, let $p=(p^i)_{i=1}^d$ and $q=(q^i)_{i=1}^d$ be frequency histograms with $d$ bins (discrete distributions called multinomials), then $J(p,q)\rightarrow \infty$ if there exists one bin $l\in\{1, ..., d\}$ such that $p^l$ is above some constant, and $q^l\to 0$. In that case, $p^l\log\frac{p^l}{q^l}\to \infty$. 
To circumvent this unboundedness problem,  the
Jensen-Shannon divergence was introduced in~\cite{Jensen-Shannon-divergence}.
The Jensen-Shannon divergence symmetrizes the Kullback-Leibler divergence by taking the {\it average relative entropy of the source distributions to the entropy of the average distribution} $\frac{p+q}{2}$:

\begin{eqnarray}
\JS(p,q) & = &\frac{1}{2}\left(\KL\left(p : \frac{p+q}{2}\right)+\KL\left(q : \frac{p+q}{2}\right)\right)=\JS(q,p) \\
    &=& \frac{1}{2} \left( H\left(p : \frac{p+q}{2}\right) - H(p)  + H\left(q : \frac{p+q}{2}\right) -H(q) \right), \\
      &= & \frac{1}{2} \int \left(p(x)\log \frac{2p(x)}{p(x)+q(x)} +  q(x)\log \frac{2q(x)}{p(x)+q(x)}  \right) \dx, \label{eq:introk}\\ 
 & = & H\left(\frac{p+q}{2} \right)  - \frac{H(p)+H(q)}{2} \geq 0.
\end{eqnarray}

The Jensen-Shannon divergence has always finite values, and its square root yields a metric, satisfying the triangular inequality.
Moreover, we have the following information-theoretic inequality~\cite{Jensen-Shannon-divergence} 
\begin{equation}
0 \leq \JS(p,q) \leq \frac{1}{4} J(p,q).
\end{equation}

By introducing the $K$-divergence~\cite{Jensen-Shannon-divergence} (see Eq.~\ref{eq:symJ}):
\begin{equation}\label{eq:khalf}
K(p:q)=\int p(x)\log \frac{2p(x)}{p(x)+q(x)} \dx = \KL\left(p : \frac{p+q}{2}\right), 
\end{equation}

we interpret the Jensen-Shannon divergence as the Jeffreys symmetrization of the $K$-divergence (see Eq.~\ref{eq:symJ}). 

\begin{eqnarray}
\JS(p,q) & = &\frac{1}{2}(K(p:q)+K(q:p)),\\
 & = & H\left(\frac{p+q}{2}\right) - \frac{H(p)+H(q)}{2}. 
\end{eqnarray}

%
%
%

Consider the {\it skewed} $K$-divergence
\begin{equation}
K_\alpha(p:q)=\int p(x)\log \frac{p(x)}{(1-\alpha)p(x)+\alpha q(x)} \dx,
\end{equation}
 and its symmetrized divergence
\begin{equation}
\JS_\alpha(p,q)=\frac{K_\alpha(p:q)+K_\alpha(q:p)}{2} =\JS_\alpha(q,p) .
\end{equation}
For $\alpha=\frac{1}{2}$, we find the Jensen-Shannon divergence: $\JS(p,q)=\JS_{\frac{1}{2}}(p,q)$.
For $\alpha=1$, we obtain half of Jeffreys divergence: $\JS_{1}(p,q)=\frac{1}{2}J(p,q)$.
It turns out that this family of  {\it $\alpha$-Jensen-Shannon divergence} belongs to a broader family of information-theoretic measures, called Ali-Silvey-Csisz\'ar divergences~\cite{Csiszar-1967,AliSilvey-1966}.
A $\phi$-divergence is defined for a strictly convex function $\phi$ such that $\phi(1)=0$ as: 

\begin{equation}
I_\phi(p:q) = \int q(x) \phi\left( \frac{p(x)}{q(x)} \right) \dx.
\end{equation}
We can always symmetrize $\phi$-divergences by taking the {\it coupled} convex function $\phi^{*}(x)=x\phi(\frac{1}{x})$.
Indeed, we get

\begin{eqnarray}
I_{\phi^*}(p:q) & = & \int q(x) \phi^*\left( \frac{p(x)}{q(x)} \right) \dx,\\
 & = & \int q(x) \frac{p(x)}{q(x)}\phi\left( \frac{q(x)}{p(x)} \right) \dx ,\\
 & = &  \int p(x) \phi\left( \frac{q(x)}{p(x)} \right)\dx = I_\phi(q:p).
\end{eqnarray}
Therefore, $I_{\phi+\phi^*}(p,q)$ is a symmetric divergence. Let $\phi^s=\phi+\phi^*$ denote the symmetrized generator.
Jeffreys divergence is a $\phi$-divergence for $\phi(u)=-\log u$ (and $\phi^s(u)=(u-1)\log u$).
Similarly, Jensen-Shannon divergence is interpreted as $\JS(p,q)=\frac{1}{2} (K(p:q)+K(q:p))$, with
$\frac{1}{2}K(p:q)$ a $\phi$-divergence for $\phi(u)=\frac{u}{2}\log\frac{2u}{1+u}$, see~\cite{Jensen-Shannon-divergence}.
It follows that Jensen-Shannon is also a $\phi$-divergence.
The $\alpha$-Jensen-Shannon divergences are $\phi$-divergences for the generators $\phi^s_\alpha=\phi^*_\alpha+\phi_\alpha$, with $\phi^*_\alpha(x)=-\log ((1-\alpha)+\alpha x)$ and $\phi_\alpha(x)=-x\log ((1-\alpha)+\frac{\alpha}{x})$.
$\alpha$-Jensen-Shannon divergences are convex statistical distances in both arguments.

One drawback for estimating $\alpha$-JS divergences on {\it continuous} parametric densities (say, Gaussian distributions), is that the mixture of two Gaussians is not anymore a Gaussian, and therefore the average distribution falls outside the family of considered distributions. 
This explains the lack of closed-form solution for computing the Jensen-Shannon divergence on Gaussians.

Next, we introduce a novel family of symmetrized divergences which yields closed-form formulas for statistical distances of a large class of parametric distributions, called statistical exponential families.

\section{A novel parametric family of Jensen divergences}

At the heart of many statistical distances lies the celebrated Jensen's convex inequality~\cite{Jensen-1906}.
For a strictly convex function $F$ and a parameter $\alpha\in\mathbb{R}\backslash\{0,1\}$, let us define the
{\it $\alpha$-skew Jensen divergence} as

\begin{equation}
J_F^{(\alpha)}(p:q) = \frac{1}{\alpha(1-\alpha)}  \int ((1-\alpha) F(p(x))+\alpha F(q(x)) - F((1-\alpha)p(x)+\alpha q(x)) \dx. 
\end{equation}

This statistical divergence is  said {\it separable} as it can be rewritten as

\begin{equation}
J_F^{(\alpha)}(p:q) = \int   j_F^{(\alpha)}(p(x):q(x)) \dx,
\end{equation}
with the scalar basic distance being defined as

\begin{equation}
j_F^{(\alpha)}(x:y) = \frac{1}{\alpha(1-\alpha)} \left( (1-\alpha) F(x)+\alpha F(y) - F((1-\alpha)x+\alpha y) \right).
\end{equation}

Furthermore, we refine the definition of $\alpha$-skew Jensen divergence to real-valued $d$-dimensional vectors $p$ and $q$ as

\begin{equation}
J_F^{(\alpha)}(p:q) = \sum_{i=1}^d    j_F^{(\alpha)}(p^i:q^i) \dx.
\end{equation}

In the limit cases, we find the oriented Kullback-Leibler divergences~\cite{2010-brbhat} when we choose generator $F(x)=-x\log x$: 

\begin{eqnarray}
\lim_{\alpha\to 0} J_F^{(\alpha)}(p:q) &= & \KL(p:q), \\
\lim_{\alpha\to 1} J_F^{(\alpha)}(p:q) &= & \KL(q:p).
\end{eqnarray}

Observe also that $J_F^{(\alpha)}(q:p)= J_F^{(1-\alpha)}(p:q)$, and that therefore  $\alpha$-skew Jensen divergences are asymmetric distortion measures (except for $\alpha=\frac{1}{2}$).
Therefore, let us  symmetrize those $\alpha$-skew divergences by averaging the two orientations as follows:

\begin{eqnarray}
\sJ_F^{(\alpha)}(p,q) & = & \frac{1}{2} (J_F^{(\alpha)}(p:q)+J_F^{(\alpha)}(q:p)) \\
& = & \frac{1}{2} (J_F^{(\alpha)}(p:q)+J_F^{(1-\alpha)}(p:q)) \\
 &= & \frac{1}{2\alpha(1-\alpha)} \int \left( F(p(x))+F(q(x)) \right. \nonumber\\ && \left. - F(\alpha p(x) + (1-\alpha) q(x)) - F( (1-\alpha) p(x) + \alpha q(x)
 \right) \dx \\
 & = & \sJ_F^{(\alpha)}(q,p) = \sJ_F^{(1-\alpha)}(p,q) \geq 0
\end{eqnarray}

For discrete $d$-dimensional parameter vectors $p$ and $q$ (with respective coordinates $p^{1}, ..., p^{d}$ and $q^{1}, ..., q^{d}$), we define analogously the $\alpha$-skew divergences as follows:

\begin{equation}
\sJ_F^{(\alpha)}(p,q) =\frac{1}{2\alpha(1-\alpha)} \sum_{i=1}^d \left( F(p^{i})+F(q^{i}) 
 - F(\alpha p^{i} + (1-\alpha) q^{i}) - F( (1-\alpha) p^{i} + \alpha q^{i}
 \right) \dx.
\end{equation}
 
Those statistical divergences are {\it separable} 
  as they can be written as 
  $\sJ_F^{(\alpha)}(p,q)=\int \sj_F^{(\alpha)}(p(x):q(x))\dx$, where $\sj_F$ denotes the corresponding basic scalar distance measure:
  
\begin{equation}
\sj_F^{(\alpha)}(x,y)= \frac{1}{2\alpha(1-\alpha)}   \left( F(x)+F(y) 
 - F(\alpha x + (1-\alpha) y) - F( (1-\alpha) x + \alpha y
 \right) \dx.
\end{equation}

\begin{figure}
\centering

\includegraphics[width=0.7\textwidth]{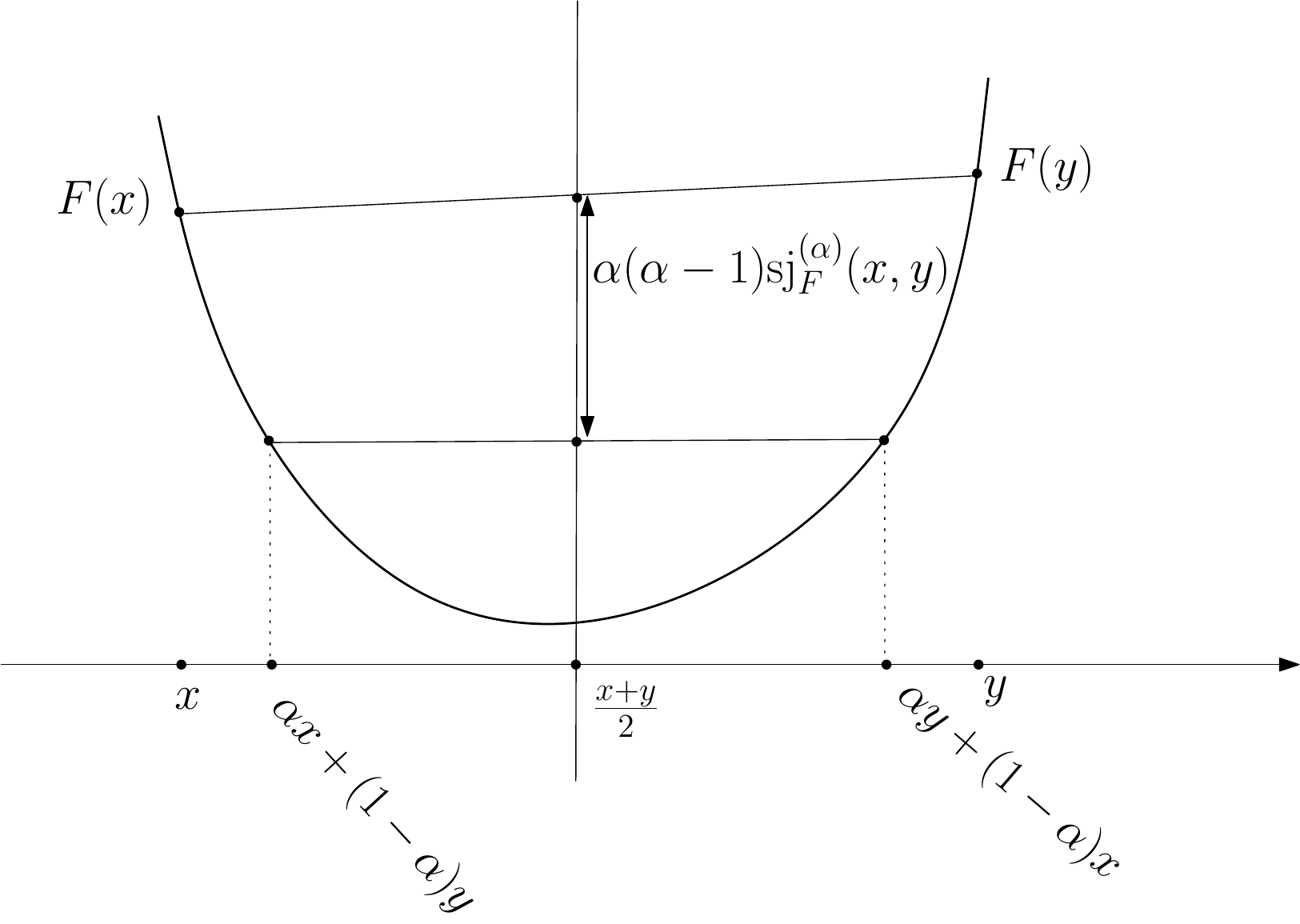}

\caption{A family of separable symmetric scalar Jensen divergences $\{\sJ_F^{(\alpha)}(p,q)=\int \sj_F^{(\alpha)}(p(x),q(x))\dx\}_\alpha$ for $\alpha\in (0,\frac{1}{2}]$ based on Jensen's convexity gap that includes both Jeffreys divergence in the limit case $\alpha=0$ and Jensen-Shannon divergence for $\alpha=\frac{1}{2}$, for the Shannon information generator. Here, we plot the scalar base distance $\sj_F^{(\alpha)}$ that induces the statistical distance.
}\label{fig:symmetric}
\end{figure}

Figure~\ref{fig:symmetric} shows graphically this novel family of symmetric Jensen divergences by depicting its associated basic scalar distance $\mathrm{sj}_F^{(\alpha)}$ (it is enough to consider $\alpha\in [0,\frac{1}{2}]$).
Note that except for $\alpha\in\{0,1\}$, this family of divergences have necessarily the boundedness property: $0\leq \sJ_F^{(\alpha)}(p,q)<\infty, \forall \alpha\not\in\{0,1\}$

Consider the strict convex generator $F(x)=x\log x$ (Shannon information function or equivalently the negative Shannon entropy functional).
Rewriting the divergence for this generator, we get a family of {\it symmetric Kullback-Leibler divergences}:

\begin{equation}
\sKL^{(\alpha)}(p,q)  = \frac{1}{2\alpha(1-\alpha)}   \left(   H(\alpha p + (1-\alpha) q) + H( (1-\alpha) p + \alpha q) - (H(p)+H(q)) 
\right) \geq 0
\end{equation}

We have in the limit case:
\begin{equation}
\lim_{\alpha\to 0} \sKL^{(\alpha)}(p,q) = J(p,q) = \sKL^{(0)}(p,q).
\end{equation}
That is, symmetrized $\alpha$-Jensen divergences tend asymptotically to the Jeffreys divergence for the Shannon information generator.
Furthermore, consider the case $\alpha=\frac{1}{2}$:

\begin{equation}
\sKL^{(\frac{1}{2})}(p,q) = 2 \left( 2 H\left(\frac{p+q}{2}\right) -  (H(p)+H(q))  \right) =4\JS(p,q).
\end{equation}
Thus this family of symmetric Kullback-Leibler divergences unify both Jensen-Shannon divergence (up to a constant factor for $\alpha=\frac{1}{2}$) with Jeffreys divergence ($\alpha\to 0$).

\newtheorem{theorem}{Theorem}
\begin{theorem}
There exists a parametric family of symmetric information-theoretic divergences $\{\sKL^{(\alpha)}\}_\alpha$ that unifies 
both Jeffreys $J$-divergence ($\alpha\rightarrow 0$) with Jensen-Shannon divergence ($\alpha=\frac{1}{2}$).
\end{theorem}

This result can be obtained by considering skew average of distributions instead of the one-half of Eq.~\ref{eq:khalf}:

\begin{eqnarray}
L_\alpha(p:q) &=& \frac{H((1-\alpha)p+\alpha q)-H(p)}{\alpha(1-\alpha)} \geq 0 
\end{eqnarray}
Then it comes out  that (see Eq.~\ref{eq:symJ})

\begin{equation}
\sKL^{(\alpha)}(p,q)  = \frac{1}{2\alpha(1-\alpha)} ( L_\alpha(p:q)  +  L_\alpha(q:p)  ).
\end{equation}
Note that $L_{\frac{1}{2}}(p:q)=4K(p:q)$. The scaling factor is due to historical convention.
However $L_\alpha$ is in general not a $\phi$-divergence. 

An alternative description of the symmetric family is given by

\begin{equation}
S_F^{(\alpha)}(p,q) = \frac{2}{1-\alpha^2}  \left(F(p) + F(q) - F\left(\frac{1-\alpha}{2}p +  \frac{1+\alpha}{2}q\right)
- F\left(\frac{1+\alpha}{2}p +  \frac{1-\alpha}{2}q\right)  \right).
\end{equation}

It can be checked that $\sJ_F^{(\alpha)}(p,q)=S_F^{(\alpha')}(p,q)$ for $\alpha'=1-2\alpha$.

Many parametric distributions follow a regular structure called exponential families. 
In the next section, we shall link that class of symmetric $\sJ^{\alpha}$-divergences to equivalent symmetric $\alpha$-Bhattacharrya divergences computed on the distribution parameter space.
 
\section{Case of statistical exponential families}

Many common statistical distributions are handled in the unified framework of exponential families~\cite{2009-BregmanCentroids-TIT,2010-brbhat}.
A distribution is said to belong to an exponential family $E_F$, if its {\it parametric} density can be canonically rewritten as

\begin{equation}
p_F(x;\theta) = \exp (\innerproduct{t(x)}{\theta}-F(\theta)+k(x)),
\end{equation}
where $\theta$ describes the member of the exponential family $E_F=\{p_F(x;\theta)\ | \theta\in\Theta\}$, characterized by the log-normalizer $F(\theta)$, a convex differentiable function. $\innerproduct{x}{y}$ denotes the inner-product (e.g., $x^Ty$ for vectors, see~\cite{2009-BregmanCentroids-TIT,2010-brbhat}). $t(x)$ is the sufficient statistic.

Discrete $d$-dimensional distributions (corresponding to frequency histograms with $d$ non-empty bins in visual applications) are multinomials, an exponential family with the dimension of the natural space $\Theta$ being $d-1$ (the order of the family).
In information retrieval~\cite{diveval:2001}, one often needs to perform clustering on frequency histograms for building a codebook to perform efficiently retrieval queries (eg., bag-of-words method~\cite{bow-2005}).

It is known that the Kullback-Leibler divergence of members $p\sim E_F(\theta_p)$ and $q\sim E_F(\theta_q)$ of the same exponential family $E_F$ is equivalent to a Bregman divergence on the swapped natural parameters~\cite{bregmankmeans-2005}:
\begin{equation}\label{eq:klbd}
\KL(p_F(x;\theta_p) : p_F(x;\theta_q)) = B_F(\theta_q:\theta_p),
\end{equation}
where $B_F(\theta_q:\theta_p) =  F(\theta_q) - F(\theta_p) - \langle \theta_q-\theta_p, \nabla F(\theta_p) \rangle$.


The Jeffreys $J$-divergence on members of the same exponential family (lhs.) can be computed as a symmetrized Bregman divergence, yielding an equivalent  calculation on the natural parameter space (rhs.):
\begin{equation}
J(p_F(x;\theta_p) , p_F(x;\theta_q)) =  (\theta_p -\theta_q)^T (\nabla F(\theta_p)-\nabla F(\theta_q))
\end{equation}

Note that although the product of two exponential families is an exponential family, it is {\it not} the case for the mixture of two exponential families.
Indeed, the mixture $(1-\alpha)p+\alpha q$ does not in general belong to $E_F$.
Therefore, the Jensen-Shannon divergence on members of the same exponential family {\it cannot} be computed directly from the natural parameters, since it requires to compute the entropy of the mixture distribution (with no known generic closed form):

\begin{equation}
\JS(p=p_F(x;\theta_p) , q=p_F(x;\theta_q))  =   H\left(\frac{p+q}{2}\right) - \frac{H(p)+H(q)}{2},\\
\end{equation}

In fact, it turns out that Eq.~\ref{eq:klbd} is the limit case of the property that $\alpha$-skew Bhattacharrya divergence $B^{(\alpha)}$ of members 
$p=p_F(x;\theta_p)$ and $q=p_F(x;\theta_q)$
 of the same exponential family $E_F$ is equivalent to a $\alpha$-Jensen divergence defined on the natural parameters~\cite{2010-brbhat}:

\begin{eqnarray}
B^{(\alpha)}(p_F(x;\theta_p) : p_F(x;\theta_q)) &=& -\log \int p_F(x;\theta_p)^{\alpha} p_F(x;\theta_q)^{1-\alpha} \dx,\\
 & = & J_F^{(\alpha)}(\theta_p:\theta_q),
\end{eqnarray}
with the $\alpha$-Jensen divergence  defined on the distribution $d$-dimensional parameter vectors as

\begin{equation}
J_F^{(\alpha)}(\theta_p:\theta_q) = \frac{1}{\alpha(1-\alpha)} \sum_{i=1}^d  ((1-\alpha) F(\theta_p^{i})+\alpha F(\theta_q^{i}) - F((1-\alpha)\theta_p^{i}+\alpha \theta_q^{i}).
\end{equation}

We can therefore symmetrize $\alpha$-skew Bhattacharrya divergences:

\begin{eqnarray}
\sB^{(\alpha)}(p_F(x;\theta_p) , p_F(x;\theta_q)) &=& \frac{1}{2} (B^{(\alpha)}(p_F(x;\theta_p) : p_F(x;\theta_q)) + B^{(\alpha)}(p_F(x;\theta_q) : p_F(x;\theta_p)) ),\\
& = &  -\frac{1}{2}\log \left(\int p^{\alpha}(x)q^{1-\alpha}(x) \dx\right) \left(\int p^{1-\alpha}(x)q^{\alpha}(x) \dx\right) \\
 & = & \alpha (1-\alpha) \sJ_F^{(\alpha)}(\theta_p,\theta_q),
\end{eqnarray}
and obtain equivalently a symmetrized skew Jensen divergence on the natural parameters.

\begin{theorem}
The symmetrized skew $\alpha$-Bhattacharyya divergence on members of the same exponential family is equivalent to a symmetrized skew $\alpha$-Jensen divergence defined for the log-normalizer and computed in the natural parameter space.
\end{theorem}

Let us now consider computing centroidal centers (say, for $k$-means clustering applications~\cite{bregmankmeans-2005}).

\section{Symmetrized skew $\alpha$-Jensen centroids}

Consider the discrete symmetrized $\alpha$-Jensen divergences (not any more on distributions but on $d$-dimensional parameter vectors).
In particular, we get  for separable divergences:

\begin{equation}
\sJ_F^{(\alpha)}(x,y)  = \frac{1}{2\alpha(1-\alpha)} \sum_{i=1}^d \left( F(x^i)+F(y^i)  - 
F(\alpha x^i + (1-\alpha) y^i) - F( (1-\alpha) x^i + \alpha y^i
 \right).
\end{equation}

This family of discrete measures includes the {\it extended Kullback-Leibler divergence} for unnormalized distributions by setting $F(x)=x\log x$.
The {\it barycenter} $b$ of $n$ points $p_1, ..., p_n$ is defined as the (unique) point that minimizes  the weighted average distance:

\begin{equation}\label{eq:min}
b=\arg\min_c \sum_{i=1}^n w_i\times \sJ_F^{(\alpha)}(p_i,c), 
\end{equation}
for $w=(w_1, ..., w_n)$ a normalized weight vector ($\forall i, w_i>0$ and $\sum_i w_i=1$).
In particular, choosing $w_i=\frac{1}{n}$ for all $i$ yields by convention the {\it centroid}.
Note that the multiplicative factor in the energy function of Eq.~\ref{eq:min} does not impact the minimum.
Thus we need to equivalently minimize:

\begin{equation}
\min_c E(c) = \min_c \sum_{i=1}^n w_i (F(p_i)+F(c)-F(\alpha p_i+(1-\alpha)c) - F(\alpha c+(1-\alpha) p_i) ).
\end{equation}

Removing the constant terms (i.e., independent of $c$), this amounts to minimize the following energy functional ($\sum_i w_i=1$):

\begin{equation}
\argmin_c E(c) = \argmin_c F(c) - \sum_i w_i ( F(\alpha p_i+(1-\alpha)c) + F(\alpha c+(1-\alpha) p_i) ).
\end{equation}

Since $F$ is convex, $E$ is the minimization of a sum of a convex function plus a concave function.
Therefore, we can apply the ConCave-Convex Procedure~\cite{NIPS2009-CCCPConvergence} (CCCP) that guarantees to converge to a minimum.
We thus bypass using a gradient steepest descent numerical optimization that requires to tune a learning step parameter.  

Initializing 

\begin{equation}
c_0=\sum_{i=1}^n w_i p_i
\end{equation}
 to the Euclidean barycenter, we iteratively update as follows:

\begin{equation}
\nabla F(c_{t+1}) = \sum_{i=1}^n w_i ( (1-\alpha) \nabla F(\alpha p_i+(1-\alpha)c_t)  + \alpha \nabla F(\alpha c_t+(1-\alpha) p_i ) ),
\end{equation}

That is,

\begin{equation}
c_{t+1}= \left(\nabla F\right)^{-1}\left( \sum_{i=1}^n w_i ( (1-\alpha) \nabla F(\alpha p_i+(1-\alpha)c_t)  + \alpha \nabla F(\alpha c_t+(1-\alpha) p_i ) ) \right)
\end{equation}
(Observe that since $F$ is strictly convex, its Hessian $\nabla^2 F$ is positive-definite so that the reciprocal gradient is $\nabla F^{-1}$ is well-defined, see~\cite{Rockafeller}.)

In the limit case, we get the following {\it fixed point}  equation:
\begin{equation}
c^*= (\nabla F)^{-1}\left( \sum_{i=1}^n w_i ( (1-\alpha) \nabla F(\alpha p_i+(1-\alpha)c^*)  + \alpha \nabla F(\alpha c^*+(1-\alpha) p_i ) ) \right).
\end{equation}

This rule is a quasi-arithmetic mean, and can alternatively be initialized using 
$c_0'=\nabla F^{-1}(\sum_{i=1}^n w_i \nabla F(p_i))$ instead of $c_0$.

Let us instantiate this updating rule for $\alpha=\frac{1}{2}$ and $w_i=\frac{1}{n}$ on Shannon and Burg information functions:

$$
\begin{array}{l|l}
\mbox{Shannon information $F(x)=x\log x-x$ } & \mbox{Burg information $F(x)=-\log x$ } \\
\nabla F(x)=\log x, (\nabla F)^{-1}(x)=\exp x & \nabla F(x)=-1/x, (\nabla F)^{-1}(x)=-1/x\\ \hline
c_{t+1}=\left(\prod_{i=1}^n \frac{c_t+p_i}{2} \right)^{\frac{1}{n}}  & c_{t+1}=\frac{n}{\sum_{i=1}^n \frac{2}{c_t+p_i} }\\
\rightarrow \mbox{Geometric update}   & \rightarrow  \mbox{Harmonic update}
\end{array}
$$

Note that for Jeffreys ($\alpha=0$) and Jensen-Shannon ($\alpha=\frac{1}{2}$) divergences, the energy function is convex, and therefore the minimum is necessarily {\it unique}. (In fact, both Jeffreys and Jensen-Shannon are two instances of the class of convex Ali-Silvey-Csisz\'ar divergences~\cite{Csiszar-1967,AliSilvey-1966}.)

Since $\alpha$-JS divergences are $\phi$-divergences (convex in both arguments), the barycenter with respect to $\alpha$-JS is unique, and can be computed alternatively using any convex optimization technique. 
Ben-Tal et al.~\cite{entropicmeans-1989} called those center points  entropic means;
They consider scalar values that can be extended to dimension-wise separable divergences, but {\it not} to normalized nor continuous distributions.

\begin{theorem}
The centroid of  members of the same exponential family with respect to the symmetrized $\alpha$-Bhattacharyya divergence can be computed equivalently as the centroid of their natural parameters with respect to the  symmetrized $\alpha$-Jensen divergence using the concave-convex procedure.
\end{theorem}

Note that for members of the same exponential family, both $c_0$ or $c_0'$ initializations are interpreted as left-sided or right-sided Kullback-Leibler centroids~\cite{2009-BregmanCentroids-TIT}.

\section{Experimental results}

Statistical distances play important roles either in {\it supervised} classification tasks (e.g., interclass distance measure for feature subset selection methods~\cite{FSS}) 
or in {\it unsupervised} clustering (e.g., centroid-based $k$-means~\cite{bregmankmeans-2005}).
%
In unsupervised settings, statistical divergences are both used in the clustering preprocessing stage for building a codebook (e.g., using $k$-means algorithm~\cite{bregmankmeans-2005}), and when answering on-line queries (e.g., classification using the nearest neighbor rule).

\begin{figure}
\centering

\includegraphics[width=9cm]{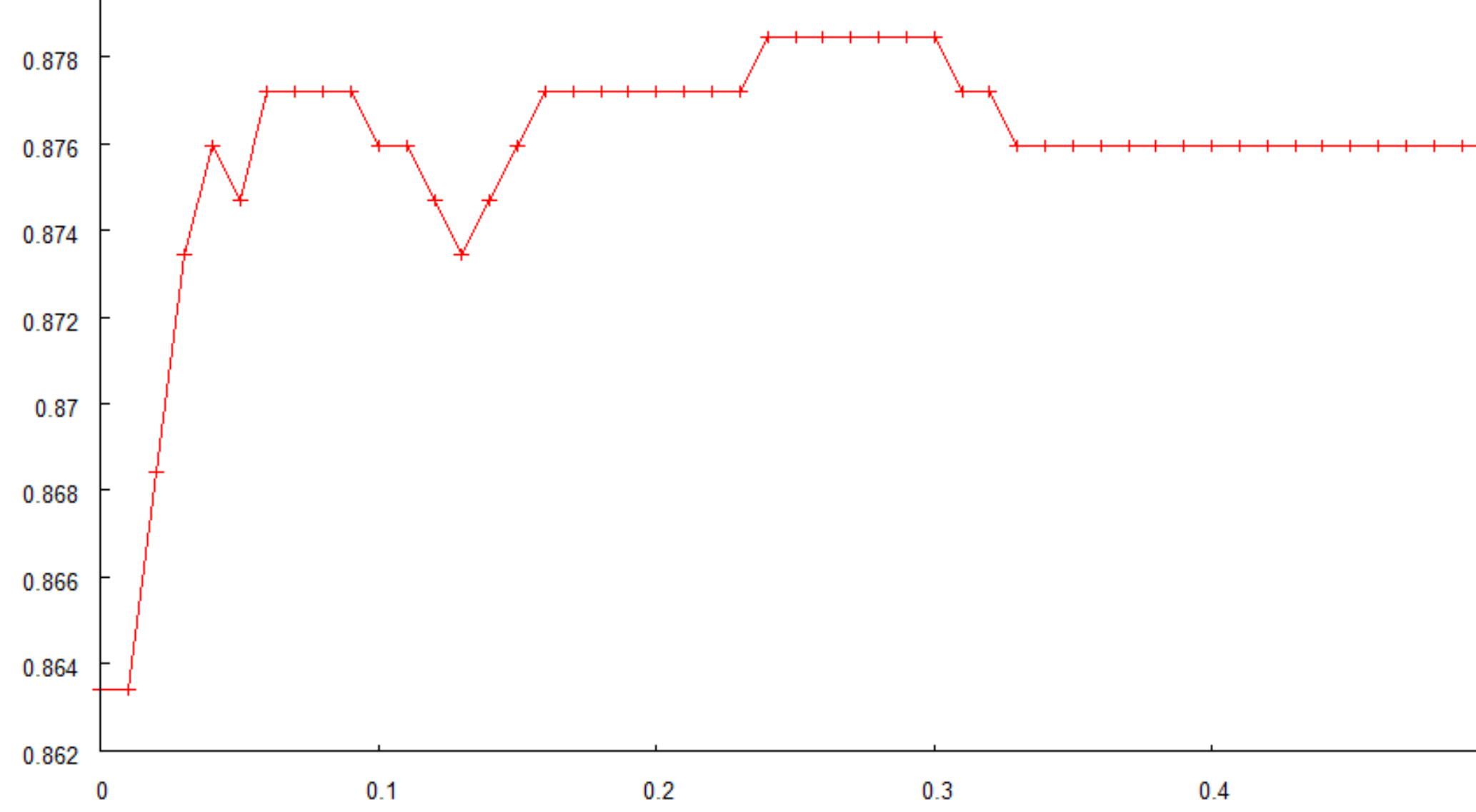}

\caption{Performance of the symmetrized $\alpha$-skew Jensen divergences for a binary classification task ($y$-axis, percentage of correct classification) with respect to $\alpha\in [0,\frac{1}{2}]$ ($x$-axis).
\label{fig:perf}}
\end{figure}

Since we proposed a novel parametric family of statistical symmetric divergences linking {\it continuously} the Jeffreys divergence to the Jensen-Shannon divergence, let us study the impact of the $\alpha$ parameter in a toy application.
Namely, we consider binary classification of images: That is, given a set of annotated images either with tag $1$ or tag $2$, perform classification of images using the nearest neighbor rule.
We use the Caltech 101 database~\cite{bow-2005} that consists of $101$ categories with about $40$ to $800$ images per category, and select the {\tt airplanes}   (800 images) and {\tt Faces} (436 images) categories.
For each color image, we choose the  intensity histogram\footnote{We convert $(R,G,B)$ colors into corresponding intensities $I=0.3R+0.596G+0.11B$, and ensure (by adding a small non-zero constant) that histogram bins are never empty in order to have proper multinomial distributions.} as its feature vector.
We compute a centroid for each of the {\tt airplane}/{\tt Faces} categories, and classify all images according to the nearest neighbor rule between those two class histogram centroids and the query image histogram with respect to the symmetrized $\alpha$-skew Jensen divergence.
Figure~\ref{fig:perf} displays the performance plot (correct percentage of classification) of this   binary classification task.
We empirically observe that performance may vary with $\alpha$ as expected, and that the best $\alpha$ needs to be tuned according to the training data hinting at the underlying geometry of data-sets. 
Here, the best correct classification rate (about $88\%$) is obtained for $\alpha=\frac{1}{4}$,
that is for the mid-divergence between Jeffreys and Jensen-Shannon divergences.
 
\section{Concluding remarks}

In this paper, we have introduced a novel parametric family of symmetric divergences based on Jensen's inequality called 
{\it symmetrized $\alpha$-skew Jensen divergences}.
Instantiating this family for the Shannon information generator, we have exhibited a one-parameter family of symmetrized Kullback-Leibler divergences.
Furthermore, we showed that for distributions belonging to the {\it same} exponential family, the symmetrized $\alpha$-Bhattacharyya divergence amounts to compute a  symmetrized $\alpha$-Jensen divergence defined on the parameter space, thus yielding a closed-form formula.
We then reported an iterative algorithm for computing the centroid with respect to this class of divergences.

For applications like information retrieval requiring symmetric statistical distances, 
the choice is therefore not anymore  to decide between Jeffreys or Jensen-Shannon divergences,  but rather to choose or tune the best $\alpha$ parameter according to the application and input data.


\section*{Acknowledgments}
The author would like to thank the anonymous reviewer for his/her valuable comments and suggestions.


\end{document}